%% file: root.tex
\documentclass[letterpaper, 10 pt, conference]{ieeeconf}  

\IEEEoverridecommandlockouts                              
\usepackage{graphicx}
\usepackage{amsmath,amsfonts}
\usepackage{romannum}
\usepackage{subcaption}
\usepackage{multirow}
\usepackage{floatrow}
\newfloatcommand{capbtabbox}{table}[][\FBwidth]
\usepackage[export]{adjustbox}
\usepackage[skip=0pt]{caption}

\title{\LARGE \bf
Deep Functional Predictive Control for Strawberry Cluster Manipulation using Tactile Prediction
}

\author{Kiyanoush Nazari${^1}$, Gabriele Gandolfi${^2}$, Zeynab Talebpour, Vishnu Rajendran${^3}$, \\ Paolo Rocco${^2}$ and Amir Ghalamzan E.${^3}$
\thanks{{$^1$} School of Computer Science,
        University of Lincoln, UK.}
\thanks{{$^2$} DEIB, Politecnico di Milano, Italy}
\thanks{{$^3$} Lincoln Institute for Agri-food Technology, University of Lincoln, UK
        }%
}

\begin{document}

\maketitle
\thispagestyle{empty}
\pagestyle{empty}

\begin{abstract}
This paper introduces a novel approach to address the problem of Physical Robot Interaction (PRI) during robot pushing tasks. The approach uses a data-driven forward model based on tactile predictions to inform the controller about potential future movements of the object being pushed, such as a strawberry stem, using a robot tactile finger. The model is integrated into a Deep Functional Predictive Control (d-FPC) system to control the displacement of the stem on the tactile finger during pushes. Pushing an object with a robot finger along a desired trajectory in 3D is a highly nonlinear and complex physical robot interaction, especially when the object is not stably grasped. The proposed approach controls the stem movements on the tactile finger in a prediction horizon. The effectiveness of the proposed FPC is demonstrated in a series of tests involving a real robot pushing a strawberry in a cluster. The results indicate that the d-FPC controller can successfully control PRI in robotic manipulation tasks beyond the handling of strawberries. The proposed approach offers a promising direction for addressing the challenging PRI problem in robotic manipulation tasks. Future work will explore the generalisation of the approach to other objects and tasks.

\end{abstract}


\section{Introduction}
\label{sec:intro}
In the field of Physical Robot Interaction (PRI), successful manipulation tasks rely on accurate interaction models that utilise rich sensory information and intelligent control strategies~\cite{billard2019trends}. Tactile feedback is a particularly effective sensing modality for PRI tasks, especially when vision-based control, such as visual servoing~\cite{mehta2016robust}, is not feasible due to occlusion~\cite{yamaguchi2019recent}. For example, pushing a ripe strawberry that is occluded by plant stems, leaves, or  unripe fruits in a cluster~\cite{xiong2019development} can require tactile feedback for effective control.


\begin{figure}[t]
        \centering
        {\includegraphics[clip, height=0.65\textwidth, trim={0cm 0.1cm 0.0cm 0cm}]{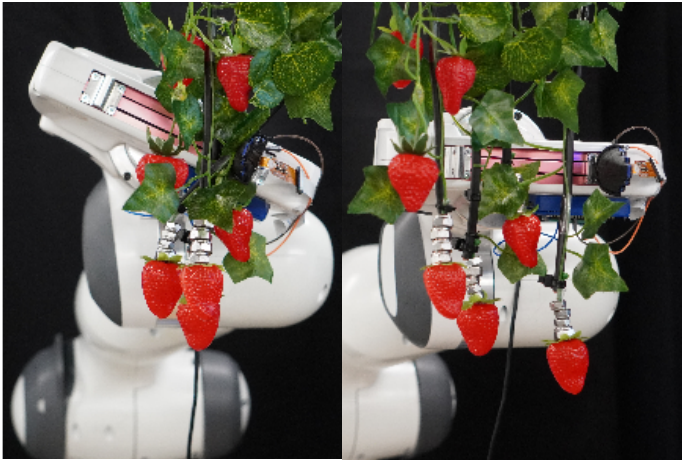}}
        \caption{Strawberry pushing setup: a Franka Emika robotic arm is pushing a cluster of strawberries from right to left where the nearest strawberry stem comes in contact with its tactile finger. (Right) the robot at the beginning of the pushing action, (Left) the robot, and cluster at the end of pushing. }
        \label{Fig:setup}
\end{figure}

Pushing is an important manipulation task that has many applications, including effective object manipulation under uncertainty~\cite{dogar2010push}, pre-grasp manipulation to position an object in a suitable configuration for grasping~\cite{king2013pregrasp}, and agile soccer ball pushing by a mobile robot ~\cite{emery2001behavior}. Analytical models for pushing require complete knowledge of the environment, including physical and geometric properties such as object pose, shape, friction parameters, and mass. Developing analytical models for unstructured environments characterized by high degrees of freedom, non-linearity, and stochasticity, such as the case of pushing a flexible stem to reach a strawberry, can be a challenging task~\cite{mohammed2022review}.

Most existing pushing methods are designed for 2-D scenarios in which an object is moving on a flat surface, but in the case of strawberry picking, a 3-D pushing scenario is more relevant~\cite{stuber2020let}. Pushing a strawberry in a 3-dimensional space is more challenging than pushing an object on a table (i.e. a 2D problem). While interactive movement primitives [10] can be used to plan pushing actions, an accurate interaction model is crucial for effectively controlling the planned motion of the strawberry during pushing in this scenario. 


In this paper, we presented a novel deep functional predictive control pipeline for the manipulation of strawberries grown on a table. Our pipeline consists of three key modules: a deep action-conditioned Tactile Forward Model (TFM), a deep Contact Localisation Model (CLM), and an online deep Functional Predictive Control (d-FPC) to generate control actions. We collected a dataset of plastic strawberries being pushed in our lab setting to train TFM, which is the state-of-the-art tactile prediction model. We also trained CLM to calibrate our tactile sensor using a dataset of strawberry pushing. Finally, d-FPC uses real-time predictions from TFM and CLM to generate robot actions based on future error signal estimations to control the stem pose on the sensor surface. We compared our proposed functional predictive controller's performance with a PD control-based system that only uses CLM and demonstrated that the predictive system outperforms this baseline model. This study addresses the challenge of pushing flexible objects in 3D, and to the best of our knowledge, this is the first study to do so. Our results demonstrate the effectiveness of our proposed approach and pave the way for future research in the manipulation of flexible objects using deep functional predictive control.

\begin{figure*}[t!]
    \centering
    \includegraphics[width=\linewidth, trim={0 0 0 0},clip]{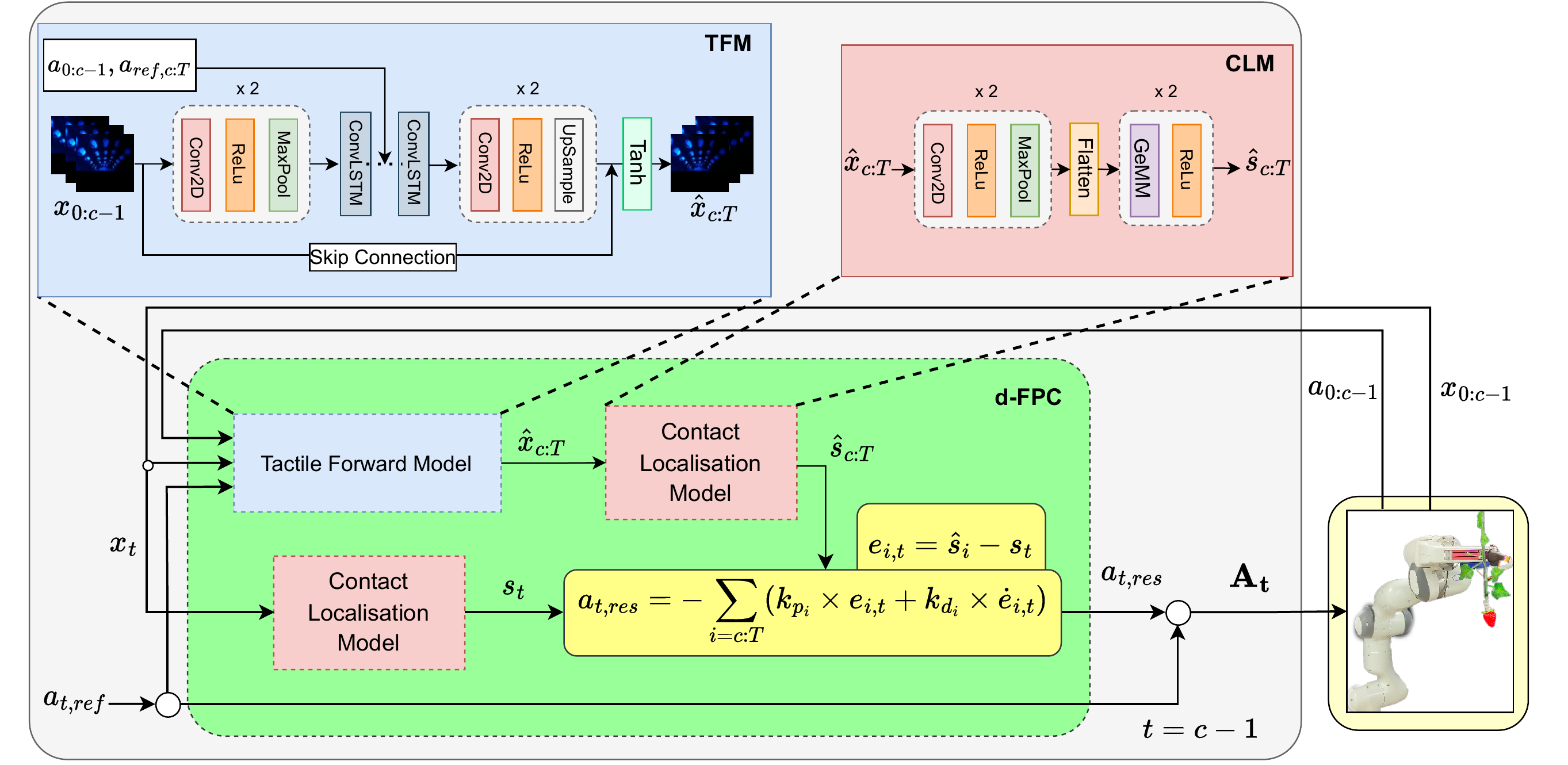}
    \label{fig:control}
 
  \caption{The block diagram of the proposed data-driven functional predictive control for pushing strawberries. The model consists of (1) tactile forward model (TFM) which is  based on~\cite{mandil2022action}, contact localisation model (CLM), and the functional predictive controller (d-FPC) that generates future actions resulting in the minimum stem displacement on the tactile finger.}
  \label{fig:controlscheme}
\end{figure*}

\begin{figure}[t]
  \begin{subfigure}[]{\textwidth}
    \centering
    \includegraphics[width=\textwidth, clip, trim={0cm 1.8cm 0.5cm 2cm}]{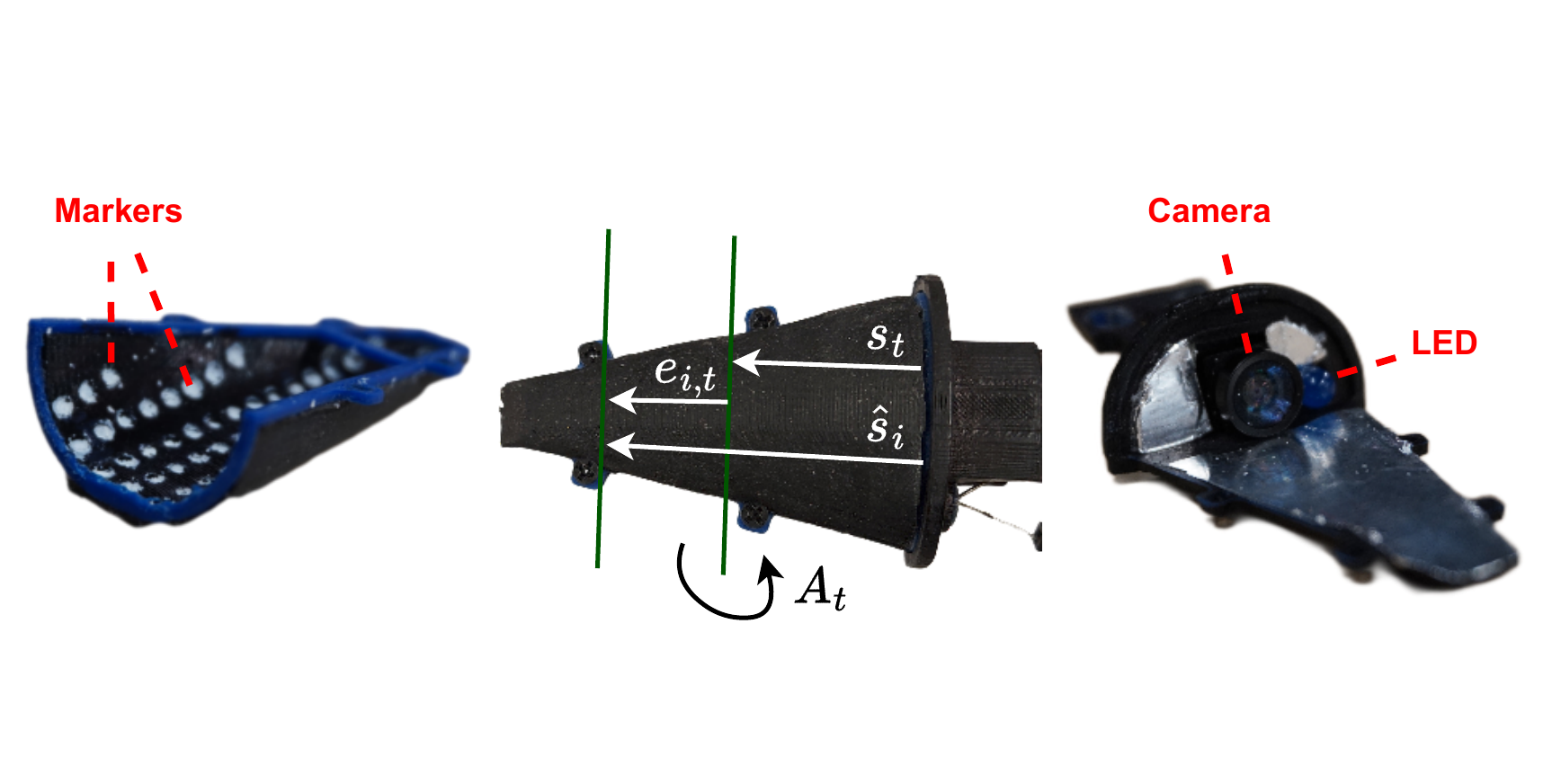}
    \caption{}
    \label{fig:finger}
  \end{subfigure}
  \begin{tabular}[c]{@{}c@{}}
    \begin{subfigure}[c]{0.32\textwidth}
      \centering
      \includegraphics[width=\textwidth, clip]{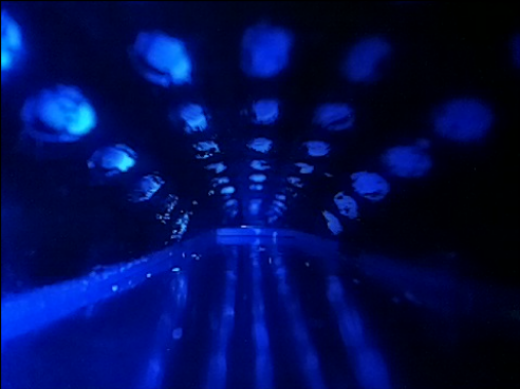}%
      \caption{}
      \label{fig:rest}
    \end{subfigure}
    \begin{subfigure}[c]{0.32\textwidth}
      \centering
      \includegraphics[width=\textwidth,page=2]{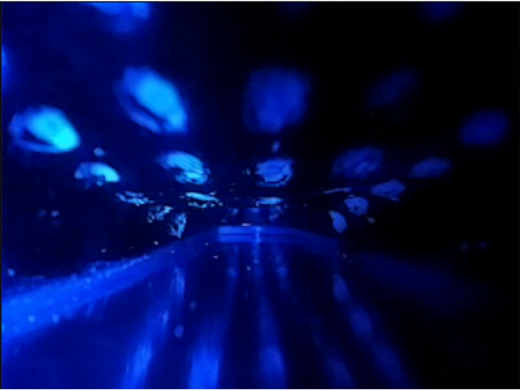}%
      \caption{}
      \label{fig:base}
    \end{subfigure}
    \begin{subfigure}[c]{0.32\textwidth}
      \centering
      \includegraphics[width=\textwidth,page=3]{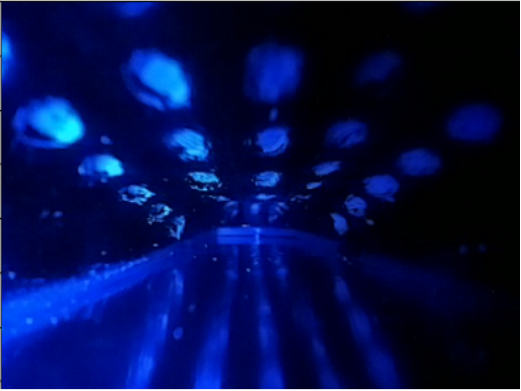}%
      \caption{}
      \label{fig:middle}
    \end{subfigure}
  \end{tabular}
  \caption{(a) Our tactile finger design features a deformable half-conic membrane with an integrated miniature camera and LED light. The initial contact line with the stem is considered as the reference line. We predict the location of the stem line $\hat{s}$ within the prediction window ${c,..., T-1}$, where $c$ denotes the context window. The robot rotates around the stem contact line to counteract the predicted stem displacement with an action $A_t$. While (b) depicts the camera readings of the tactile finger at rest, (c) and (d) show when forces are applied to the membrane near the base and between the base and middle point, respectively.
  }
  \label{fig:fig3}
\end{figure}

\section{Related works}
\label{sec:litrev}

Cluster manipulation in fruit harvesting is a challenging task from both motion planning and motion control perspectives~\cite{zhou2022intelligent,mghames2020interactive}. One of the challenges is avoiding slip of a grasped object, which can be addressed through closed-loop robot trajectory adaptation~\cite{nazari2022proactive}. Deformable object manipulation, such as cloth, has been modelled using simplified mass-spring models or 3D mesh generation~\cite{arriola2020modeling}, while heuristic feature spaces have been used for flexible cable manipulation with dual robot arms~\cite{zhu2018dual}. However, analytical modelling methods are limited to specific object sets and are not scalable to larger object and action sets. In contrast, our proposed approach uses a time-series model for action-conditioned tactile prediction for pushing control which can be applied in unstructured settings without the knowledge about the model of the individual objects.

Tactile feedback is mostly used for grasp control in robotic object clutter manipulation~\cite{yamaguchi2019recent} and detecting a grip on the fruit, detaching and dropping into a basket in harvesting settings~\cite{zhou2023branch}. Another line of research uses tactile sensors for ripeness estimation~\cite{chen2022non} and slip detection during fruit picking~\cite{dischinger2021towards}. However, the use of tactile sensors has been limited to grip control and has not been applied for any cluster manipulation. In our work, we exploit tactile feedback for trajectory-level control for pushing a flexible plant stem. 

Tactile prediction models are used for controlling manipulation tasks, from the simple task of rolling a marble on a table~\cite{tian2019manipulation} to the complex task of slip control~\cite{nazari2022proactive}. The core of such controllers is a forward model that can generate predicted tactile readings (we call them tactile images). For instance, action-conditioned tactile predictive models are utilised with a taxel-based tactile sensor in pick and place tasks~\cite{mandil2022action}, demonstrating the approach performs well only for flat surface objects. 

Our approach uses a time-series model for tactile prediction based on~\cite{mandil2022action}. We form a deep Predictive Functional Control (d-FPC)~\cite{rossiter2019new, rossiter2022recent} which enables the robot to control the strawberry pushing actions. Deep models have been extensively used for learning lower dimensional state spaces for Model Predictive Control (MPC)~\cite{lenz2015deepmpc}. These methods have also been used for learning visual dynamic models for control~\cite{nagabandi2020deep}. In a simplified task of rolling a dice, the tactile prediction was used in an MPC controller \cite{tian2019manipulation}. In our work, we form a Proportional-Derivative (PD) control over the error in the prediction horizon to control the contact state of a flexible object on a robot hand. Unlike previous work that used trajectory adaptation to minimise the likelihood of predicted binary slip signal in a prediction horizon~\cite{nazari2022proactive}, our model learns the complex contact behaviour and generates actions to control the movements of the stem on the tactile finger to keep it stable.

\section{Methodology}
\label{sec:mehtod}
\label{sec:Problem Fprmulation}
\paragraph{Camera-based tactile sensor} We use a customised camera-based tactile sensor for pushing strawberries similar to Tactip~\cite{lloyd2021goal}. This sensor has a camera and an LED light looking at a deformable membrane with embedded white markers (Fig.~\ref{fig:fig3}). The applied pressure on the sensor yields a deformation that is captured by the camera.


\paragraph{Contact Localisation Model (CLM)} The motions of the marker array printed on the sensor are indicative of the magnitude and location of the applied force. For the current problem setting, we are more interested in force localisation for doing stem contact state control. To find the mapping from raw tactile images to contact location in 1-dimensional space, we use a Convolutional Neural Network with the architecture shown in Fig.\ref{fig:controlscheme} (red box). CLM consists of two convolutional and three dense layers. The output of CLM is the distance of the contact force from the sensor camera lens along the sensor conic axis. The data set for training CLM consists of applying forces to the fixed sensor by a rod (mimicking strawberry stem) attached to the robot end-effector (EE) with a 5mm distance step. At each step, the robot applies force on the membrane toward the sensor base by a 1mm penetration step. Overall, 150 stem pushing samples in 10 locations are collected to train CLM.

\paragraph{Tactile Forward Model (TFM)} Here, we present the formulation of the tactile prediction problem for our custom-made camera-based tactile sensor. Tactile prediction aims to estimate future tactile images based on a set of previous tactile images $\textbf{x}_{0},...,\textbf{x}_{c-1}$ obtained from physical interactions, where $c$ is the length of the context window. Specifically, the objective is to sample from the conditional distribution $p(\textbf{x}_{c:T}|\textbf{x}_{0:c-1})$, where $\textbf{x}_{i}$ denotes the i$^{th}$ tactile image in the sequence and $T$ is the sum of the context window length and the prediction horizon length.

Since the robot's actions alter the environment during physical interaction, we incorporate action conditioning to predict tactile sensation more accurately. The action-conditioned tactile prediction problem is formulated as predicting the future tactile images $\textbf{x}_{c:T}$ given a sequence of previous robot actions $\textbf{a}_{0:c-1}$, previous tactile images $\textbf{x}_{0:c-1}$, and a sequence of future/planned robot actions/trajectory $\textbf{a}_{c:T}$. Here, a robot action, $\textbf{a}\in\mathbb{R}^{6}$, refers to the end-effector task space position and orientation (Euler angles) with respect to the robot base, while a tactile image is represented by $\textbf{x} \in\mathbb{R}^{64 \times 64 \times 3}$, which captures the surface deformation caused by the applied force. The conditional distribution will be:

\begin{equation}
    p(\textbf{x}_{c:T}|\textbf{x}_{0:c-1}, \textbf{a}_{0:T})   
\end{equation}
Factorising this we can define the model as $\Pi^{T}_{t=c}p_{\theta}(\textbf{x}_{t}|\textbf{x}_{0:t-1}, \textbf{a}_{0:t})$. Learning now involves training the parameters of the factors $\theta$.

The model architecture is depicted in Fig.\ref{fig:controlscheme} (blue box). We extract scene features from the input tactile image by convolutional filters in the first two layers of the network as the encoder. Each convolution operation is followed by the Relu activation function and 2D maxpooling operations. Robot action sequences are concatenated with latent tactile features after the convolutional layers. These latent space features with downsampled width and height and a larger number of channels are fed to the Conv-LSTM chain. These layers process the spatiotemporal dependencies among the latent features. After this point, we need to upscale the features to reach the tactile image size. As such, two convolutional layers, each one followed by Relu activation and 2D upsampling, are applied to ConvLSTM outputs. To apply the pixel motion changes to the input, we use the skip connection for the input tactile image and apply $tanh$ activation to construct the next tactile images in the sequence.

\paragraph{deep-Functional Predictive Control (d-FPC)} We denote the predicted stem location (from CLM) on the sensor at time $t$ by ${s_t}$. The goal of our d-FPC is to control the stem displacement on the tactile finger. Hence, this allows the robot to keep the contact fixed with the strawberry stem during pushing actions and avoid the contact location approaching the tip or the base of the sensor. These are sensor surface boundary zones and approaching them increases the probability of losing contact with the stem. We use the stem-finger contact point at time $t$ as the reference for our d-FPC controller. We define an error signal as the distance of the contact point from the reference point:
\begin{equation}
    e_{i, t} = \hat{s}_i - s_{t},\; i = {c, . . ., T}
    \label{eq:et}
\end{equation}
where $\hat{s}_i$ is the predicted stem location for a sequence of planned robot movements. We formulate our d-FPC over the error signal as follows:
\begin{equation}
    a_{t, res} = -\sum_{i=c:T} (k_{p_i} \times e_{i, t} + k_{d_i} \times \dot{e}_{i, t})
    \label{eq:action}
\end{equation}
where $a_{t, res}$ is the residual action value to be added to the reference trajectory $a_{t, ref}$ to generate the control action $A_t$. $A_t$ is a rotational velocity around the contact line axis. Fig.\ref{fig:controlscheme} (green box) shows the schematic of the d-FPC. The generated control output is a rotational velocity proportional to the distance of the stem from the reference line. The derivative term avoids overshooting and having large instant rotations.

\section{Experimental setup and data set}
\paragraph{Tactile sensor and manipulation task} Various types of tactile sensors are discussed in the literature, including in \cite{yamaguchi2019recent}. In this work, we use a custom-made camera-based tactile sensor based on tactip~\cite{ward2018tactip} that has a half-conic geometry and a tapered tip (shown in Fig.\ref{fig:fig3}) designed to allow for easier penetration among stems and fruits, providing valuable tactile feedback. The deformable membrane of the sensor is 3D-printed and dot features are printed with a linear pattern on its conic inner surface. Changes in the marker pattern resulting from contact forces provide information about contact force value, geometry, and location, as shown in Fig.~\ref{fig:rest}, \ref{fig:base} and \ref{fig:middle}. The camera, which is located on the sensor base, and the LED, used for illuminating the markers, are powered by an onboard Raspberry Pi, and tactile images are transmitted at a frequency of 60 Hz. The sensor is mounted on a Franka Emika gripper, providing an effective and versatile tool for physical interaction in a range of applications.

We have collected the data from a series of strawberry-pushing tasks in 3-D. The pushing dataset includes data for single strawberry pushing and pushing a cluster of strawberries. To simulate the table-top strawberry growing scenario, we attached each plastic strawberry to a thin wire that makes a nonlinear elastic behaviour similar to those usually observed in tabletop-grown strawberries. 
To simulate realistic tactile feedback, we added knots on the stalk of each strawberry (Fig.\ref{Fig:setup}) and injected silicone to increase their weight (each strawberry weighs c. 20 g to 30 g).

We generate the pushing trajectories for the training data collection phase by two methods: first by Pilz industrial motion planner by specifying initial and target robot poses, and second by defining a minimum time reference trajectory using the robot's Cartesian velocity controller. We use the second method to be able to regenerate comparably similar trajectories in test time, as opposed to the first case where trajectories are generated by the motion planning library.
Trajectories include linear and circular motion patterns to perform the pushing tasks. Arc trajectories were used to collect more tactile-conditioned robot movements, where the finger followed the motion of the pushed stem/strawberry. These pushes started at a position $p_{0}$ and orientation $q_{0}$, followed an arc trajectory, and ended at a final position $p_{f}$ with a value of $z$ coordinate larger than initial position. The final orientation $q_{f}$ is selected to maintain contact with the elements pushed. The pushing actions were performed from right to left and vice versa, and they involved single or multiple stems (Fig.~\ref{Fig:setup}), generating greater deformations on the membrane.

We collected a total of 430 mixed linear/circular motion tasks containing (\romannum{1}) tactile images from the finger at 60 Hz and (\romannum{2}) robot state data sampled at 1000 Hz, representing the position and orientation of the end effector in the planned trajectory. These readings were synchronised using the ROS~\textit{ApproximateTime} policy and fed into the tactile forward model both in training and test times.

Considering the robot's motion, slip occurred mainly on the width and length of the finger but could also happen in other directions depending on the motion of the stems during the pushing actions.

\section{Results and discussion}
\label{sec:result}


We test the performance of our proposed control pipeline in real-time on pushing tasks of strawberry stems and compare the performance with a baseline controller and an open-loop system. The tactile sensor is mounted on Franka Emika robot connected to a PC with Intel® Core™ i7-8700K CPU @ 3.70GHz × 12 and 64GB RAM running Ubuntu 20.04 and ROS Noetic. Torch library is used for offline training and online testing of the neural network models. Test manipulation tasks consist of performing pushing trajectories with linear and circular motion patterns using the robot's Cartesian velocity controller.

Performance metrics include: (I) Stem max displacement and (II) the number of stem slip instances on the sensor surface. If we denote stem location at time $t$ by $s_i$ where $i \in (0,1,...,T)$ for a pushing trial, metric (I) is defined as the absolute value of the difference of maximum and minimum stem location in a trial $|max(s_i) - min(s_i)| : i=1,...,T$. Metric (II) is defined as the number of time steps where the differential values $\dot{s}_i$ were larger than threshold $\gamma$. While metric (I) shows full stem displacement, metric (II) shows the stem's sudden large motion instances or slippage on the sensor surface. We also present the area under the curve of stem displacement and generated action.
We repeat each test case 5 times and present the mean and standard deviation of the metric values. Overall we conducted 100 test-pushing trials.

To evaluate the effectiveness of d-FPC for pushing control, we compare the control performance with a PD control-based tactile servoing system as the baseline model. Both models' results are presented against the open-loop system with a pre-specified reference trajectory.
 \input{parts/tables}
In this paper, we utilise a minimum-time reference trajectory (such as bang-bang) for the open-loop system, although any desired reference trajectory can be used. To make valid comparisons among trials, we consider three initial contact zones for the stem including \textbf{Zone-1} where the contact point is between the middle and tip of the sensor, \textbf{Zone-2} has the contact point between the middle and base of the sensor, and \textbf{Zone-3} where the contact point is close to sensor centre line. Since the tactile sensor has varying deformation limits across its conic axis we compare the trials with corresponding initial contact zones together.

We conduct a comparison test with a one-degree-of-freedom (DOF) horizontal pushing along the $Y$-axis of the robot's base frame. Both PD and d-FPC controllers generate control actions for the robot hand's rotation around the contact line to prevent stem slip on the sensor surface. The results are presented in Table~\ref{tab:control}, where test cases are conducted separately for each initial contact zone. Both PD and d-FPC controllers decrease the stem's maximum displacement.
We observe that d-FPC outperforms the PD controller for Zone-1 and Zone-3, but PD shows better performance for Zone-2 very close to the sensor base. This is because the sensor has its largest deformation limit in the Base zone, resulting in relatively large initial deformation after making contact, making it difficult for TFM to predict future stem states. The prediction of the error signal helps d-FPC to have more reaction time than PD.

We find that d-FPC is the most effective controller to reduce the number of stem slip instances, with the smallest area under the curve of displacement compared to the PD controller. We also present the computation time to show the relative computation complexity of each system. Since d-FPC has two stacked deep models, the computation time is larger than the PD controller.

\begin{figure}[tb!]
     \centering
     \begin{subfigure}[b]{.8\columnwidth}
         \centering
         \includegraphics[trim={0 0 1.1cm 1.2cm},clip,width=\textwidth]{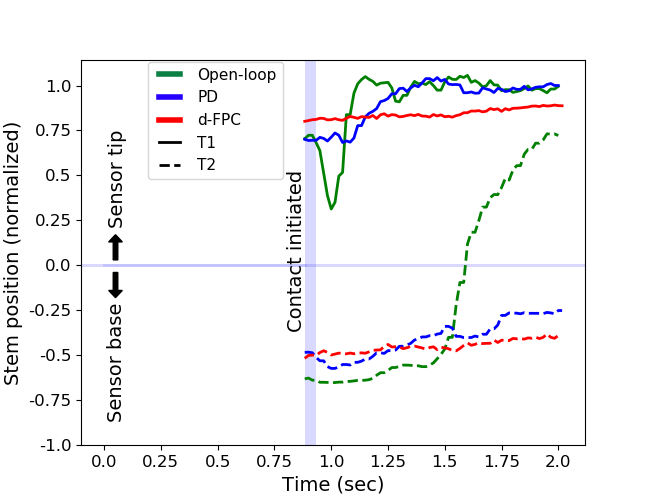}
         \caption{stem pose}
         \label{fig:stem}
     \end{subfigure}
     \\ 
     \begin{subfigure}[b]{.8\columnwidth}
         \centering
         \includegraphics[trim={0 0 1.1cm 1.2cm},clip,width=\textwidth]{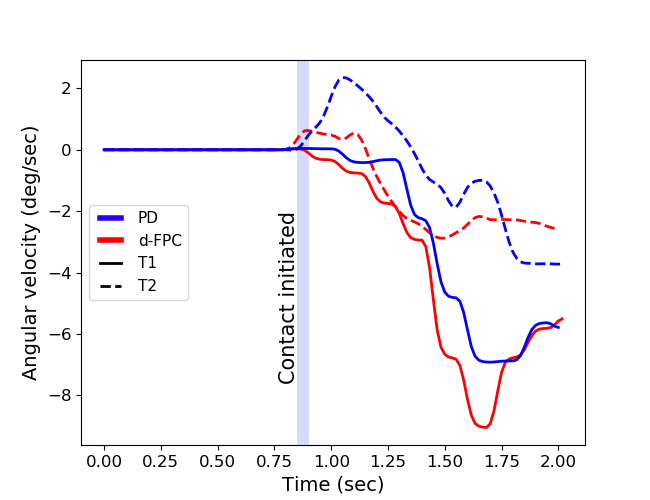}
         \caption{control action}
         \label{fig:action}
     \end{subfigure}
        \caption{Comparison of control performance between d-FPC and open loop, as well as PD controller, in maintaining the location of the stem constant on the finger surface (Trial-1 (T1) solid and Trial-2 (T2) dashed lines) (a) At time 0.85 s, the stem makes contact with the tactile finger and the controllers activate. The graph shows that d-FPC can maintain the stem contact point on the tactile finger during the pushing action, while the open loop result for the trial where the contact is closer to the sensor base shows the stem moving out of the tactile finger surface. (b) The magnitude of the control input shows d-FPC provides larger wrist rotation to avoid stem contact displacement. }
        \label{fig:stempose}
\end{figure}

To compare the performance of different controllers in a qualitative manner, we present the stem location obtained in two trials (shown in Fig. \ref{fig:stem}): Trial-1, where the stem-finger initial contact point is in Zone-1, is shown with solid lines, and Trial-2, with the contact point in Zone-2, is shown with a dashed line. Our results show that d-FPC outperforms PD controller and open loop in maintaining the stem contact, resulting in the smallest displacement of the stem. Furthermore, Fig.~\ref{fig:action} shows the control actions generated by each controller. We observe that d-FPC generates actions of larger magnitude in Trial-1 because the likelihood of losing the stem in Zone-1 (namely closer to the tip) is larger than in Zone-2. In Trial-2, the magnitude of d-FPC and PD controller actions is similar since the contact between the stem and sensor membrane is tighter due to a larger deformation of the sensor closer to the sensor base.

We test the performance of the systems in a three DOF task with a bang-bang reference for translation along $Y$, $Z$, and rotation $W_x$ of Cartesian velocity space. This is a more challenging task because the robot wrist will rotate 45 degrees along the pushing trajectory which causes larger deformation of the stem and more slip instances. Based on Table~\ref{tab:trajectory} d-FPC is the most effective controller in decreasing the stem displacement and slip instances. PD has a smaller improvement in max displacement for the circular motion than the linear motion compared to the open-loop system. This indicates that not having enough reaction time in this task can lead to failure in achieving the control objective.

We test the generalisation performance of the pushing controller when pushing a stem in a cluster of strawberries. In this task additional to the target stem, other stems, leaves, or strawberries come into contact with the sensor which makes both tactile prediction and control more challenging. Table~\ref{tab:cluster} shows the results for pushing a stem in a cluster. Although the control performance of PD and d-FPC degrades compared to pushing an isolated stem, both systems improve the performance metrics relative to the open-loop system.

 \begin{figure}[t!]
        \centering
        \includegraphics[clip, width=0.9\textwidth, trim={0.0cm 0.0cm 1cm 1cm}]{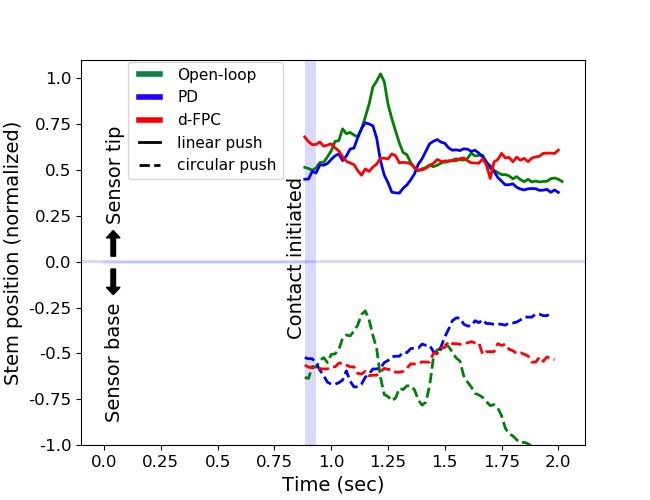}
        \caption{Strawberry cluster pushing results.}
        \label{Fig:clusterpush}
\end{figure}

Fig.\ref{Fig:clusterpush} shows cluster pushing results for sample trials of linear and circular pushing trajectories. For the linear push, PD has slight improvement compared to the open-loop system but d-FPC reduces stem displacement more effectively. For the circular push, while the open-loop system loses contact with the stem because of large stem slippage in the last part of the trial, both PD and d-FPC reduce the stem displacement to avoid large slips. d-FPC keeps the displacement more bounded relative to the PD controller does. 

\section{Conclusion }
\label{sec:discussion}
We presented a novel deep Predictive Functional Control (d-PFC) framework to control the contact location of a strawberry stem on our tactile finger. Our proposed method leverages a time-series model for generating action-conditioned tactile predictions and a convolutional neural network (CNN) model converting the tactile images to contact location. We demonstrated the effectiveness of our approach through a series of experiments with a Franka Emika robot and a customised tactile finger, showing that our model can learn complex contact behaviours and generate actions to control the movements of flexible objects to keep them stable, e.g. pushing a cluster of strawberries.

Overall, our work highlights the potential of deep learning-based approaches in addressing the challenges of tactile sensing-based manipulation tasks and lays the foundation for future research in this field.

\bibliographystyle{IEEEtran}

\bibliography{IEEEexample,egbib}

\end{document}

%% file: parts/tables.tex
%

\begin{table*}[tb!]
        \centering
        \caption{\centering Control performance for the PD and d-FPC in pushing a single strawberry along a linear trajectory.}
        \label{tab:control}
        \begin{tabular}{|c||c|c|c|c|c|c|}
            \hline
            Model & \begin{tabular}{@{}c@{}} Contact \\ zone \end{tabular} & \begin{tabular}{@{}c@{}}Stem max\\ disp. \end{tabular} & \begin{tabular}{@{}c@{}}Stem slip\\ instances \end{tabular} & \begin{tabular}{@{}c@{}}Disp. \\ integral \end{tabular} & \begin{tabular}{@{}c@{}}Action \\ integral \end{tabular} & \begin{tabular}{@{}c@{}}Comp.\\ time (ms) \end{tabular} \\\hline
            \multirow{3}{*}{\begin{tabular}{@{}c@{}}Open-loop\end{tabular}} & 1 & 0.80 $\pm$ 0.2 & 31.23 $\pm$ 4.3 & 0.83 $\pm$ 0.1 & - & - \\
            & 2 & 1.35 $\pm$ 0.2 & 50.19 $\pm$ 5.7 & 0.91 $\pm$ 0.1 & - & - \\
            & 3 & 0.91 $\pm$ 0.1 & 39.83 $\pm$ 3.2 & 0.86 $\pm$ 0.2 & - & - \\
            \hline
            \multirow{3}{*}{\begin{tabular}{@{}c@{}} PD\end{tabular}} & 1 & 0.65 $\pm$ 0.1 & 27.2 $\pm$ 6.5  & 0.75 $\pm$ 0.1 & 2.93 $\pm$ 0.7 & 18.73 $\pm$ 2 \\
             & 2 & \textbf{0.36} $\pm$ 0.0 & 10.2 $\pm$ 2.4 & 0.48 $\pm$ 0.0 & 5.12 $\pm$ 3.8 & 20.30 $\pm$ 1 \\
            & 3 & 0.63 $\pm$ 0.1 & 24.2 $\pm$ 1.6 & 0.47 $\pm$ 0.1 & 9.73 $\pm$ 5.4 & 19.73 $\pm$ 1 \\
           \hline
           \multirow{3}{*}{\begin{tabular}{@{}c@{}} \textbf{d-FPC}\end{tabular}} & 1 & \textbf{0.20} $\pm$ 0.0 & \textbf{5.0} $\pm$ 1.2  & \textbf{0.12} $\pm$ 0.0 & 3.74 $\pm$ 0.8 & 60.49 $\pm$ 6 \\
             & 2 & 0.43 $\pm$ 0.0 & \textbf{7.2} $\pm$ 0.7 & \textbf{0.18} $\pm$ 0.0 & 4.27 $\pm$ 1.2 & 55.02 $\pm$ 2 \\
            & 3 & \textbf{0.25} $\pm$ 0.1 & \textbf{6.0} $\pm$ 0.6 & \textbf{0.09} $\pm$ 0.0 & 4.57 $\pm$ 2.4 & 58.54 $\pm$ 3 \\
           \hline
        \end{tabular}
    \end{table*}
%


\begin{table*}[tb!]
        \centering
        \caption{\centering Comparison of the controllers in linear and circular pushing trajectories (* integral is the integral of the * magnitude.).}
        \label{tab:trajectory}
        \begin{tabular}{|c||c|c|c|c|c|}
            \hline
            Model & \begin{tabular}{@{}c@{}} Robot \\ trajectory \end{tabular} & \begin{tabular}{@{}c@{}}Stem max\\ disp. \end{tabular} & \begin{tabular}{@{}c@{}}Stem slip\\ instances \end{tabular} & \begin{tabular}{@{}c@{}} Disp.\\ integral \end{tabular} & \begin{tabular}{@{}c@{}} Action\\ integral \end{tabular} \\
            \hline
            \multirow{2}{*}{\begin{tabular}{@{}c@{}} Open-loop\end{tabular}} & Linear & 1.21 $\pm$ 0.18 & 44.38 $\pm$ 10.3  & 0.88 $\pm$ 0.4  & -\\
             & Circular & 1.35 $\pm$ 0.46 & 48.18 $\pm$ 5.2 & 1.02 $\pm$ 0.5 & -\\
            \hline
            \multirow{2}{*}{\begin{tabular}{@{}c@{}} PD\end{tabular}} & Linear & 0.58 $\pm$ 0.21 & 25.53 $\pm$ 4.2  & 0.63 $\pm$ 0.1 & 5.39 $\pm$ 6.2 \\
             & Circular & 1.20 $\pm$ 0.01 & 17.6 $\pm$ 2.0 & 0.44 $\pm$ 0.0 & 9.89 $\pm$ 0.8\\
           \hline
           \multirow{2}{*}{\begin{tabular}{@{}c@{}} \textbf{d-FPC}\end{tabular}} & Linear & \textbf{0.29} $\pm$ 0.04 & \textbf{8.11} $\pm$ 1.4  & \textbf{0.13} $\pm$ 0.0 &  4.49 $\pm$ 2.5\\
             & Circular & \textbf{0.54} $\pm$ 0.05 & \textbf{5.0} $\pm$ 1.5 & \textbf{0.22} $\pm$ 0.0 & 6.66 $\pm$ 0.8\\
           \hline
        \end{tabular}
    \end{table*}
%

\begin{table*}[tb!]
        \centering
        \caption{\centering Controller and open loop performances for Pushing a cluster of strawberries.}
        \label{tab:cluster}
        \begin{tabular}{|c||c|c|c|c|}
            \hline
            Model & \begin{tabular}{@{}c@{}} Robot \\ trajectory \end{tabular} & \begin{tabular}{@{}c@{}}Stem max\\ disp. \end{tabular} & \begin{tabular}{@{}c@{}}Stem slip\\ instances \end{tabular} & \begin{tabular}{@{}c@{}} Disp.\\ integral \end{tabular} \\
            \hline
            \multirow{2}{*}{\begin{tabular}{@{}c@{}} Open-loop\end{tabular}} & Linear & 1.43 $\pm$ 0.30 & 49.33 $\pm$ 15.64  & 1.39 $\pm$ 0.33 \\
             & Circular & 1.29 $\pm$ 0.67 & 47.98 $\pm$ 6.33 & 1.19 $\pm$ 0.23 \\
            \hline
            \multirow{2}{*}{\begin{tabular}{@{}c@{}} PD\end{tabular}} & Linear & 0.79 $\pm$ 0.21 & 29.4 $\pm$ 6.52  & 0.66 $\pm$ 0.21  \\
             & Circular & 1.14 $\pm$ 0.23 & 20.5 $\pm$ 2.69 & 0.56 $\pm$ 0.84 \\
           \hline
           \multirow{2}{*}{\begin{tabular}{@{}c@{}} \textbf{d-FPC}\end{tabular}} & Linear & \textbf{0.31} $\pm$ 0.08 & \textbf{17.1} $\pm$ 2.39  & 0.25 $\pm$ 0.03\\
             & Circular & \textbf{0.61} $\pm$ 0.11 & \textbf{9.5} $\pm$ 4.58 & 0.27 $\pm$ 0.18 \\
           \hline
        \end{tabular}
    \end{table*}